\documentclass[9pt, conference]{IEEEtran}
\IEEEoverridecommandlockouts
\usepackage{cite}
\usepackage{amsmath,amssymb,amsfonts}
\usepackage{graphicx}
\usepackage{textcomp}
\usepackage{xcolor}
\usepackage{svg}
\usepackage{subfig}
\usepackage{booktabs} 
\usepackage{multirow} 
\usepackage{pifont}
\usepackage{colortbl}
\usepackage{comment}

\usepackage{algorithm}
\usepackage{algpseudocode}
\usepackage{amsmath}
\usepackage{amsfonts}
\usepackage{algorithmicx}
\usepackage{algpseudocode}
\usepackage{makecell}

\definecolor{darkgreen}{rgb}{0.0, 0.5, 0.0}
\definecolor{darkred}{rgb}{0.8, 0.0, 0.0}
\definecolor{darkblue}{rgb}{0.0, 0.0, 0.8}

\usepackage[a4paper, total={184mm,239mm}]{geometry}
\def\BibTeX{{\rm B\kern-.05em{\sc i\kern-.025em b}\kern-.08em
    T\kern-.1667em\lower.7ex\hbox{E}\kern-.125emX}}

\begin{document}

\title{FPQVAR: Floating Point Quantization for Visual Autoregressive Model with FPGA Hardware Co-design}

\author{
    Renjie Wei$^{12}$,
    Songqiang Xu$^{4}$, Qingyu Guo$^2$, and Meng Li$^{123\dag}$
\\
\textit{$^1$Institute for Artificial Intelligence \& $^2$School of Integrated Circuits, Peking University, Beijing, China} \\
\textit{$^3$Beijing Advanced Innovation Center for Integrated Circuits, Beijing, China} \\
\textit{$^4$School of Software and Microelectronics, Peking University, Beijing, China}

\thanks{

$^\dag$Corresponding author: meng.li@pku.edu.cn.
Our algorithm code is available at: https://github.com/PKU-SEC-Lab/FPQVAR.}
}

\maketitle

\begin{abstract}

Visual autoregressive (VAR) modeling has marked a
paradigm shift in image generation from next-token prediction to next-scale prediction.
VAR predicts a set of tokens at each step from coarse to fine scale, leading to better image quality and faster inference speed compared to existing diffusion models.
However, the large parameter size and computation cost hinder its deployment on edge devices.
To reduce the memory and computation cost,
we propose FPQVAR, 
an efficient post-training floating-point (FP) quantization framework for
VAR featuring algorithm and hardware co-design.
At the algorithm level, 
we first identify the challenges of quantizing VAR.
To address them, we propose Dual Format
Quantization for the highly imbalanced input activation.
We further propose Group-wise Hadamard Transformation and GHT-Aware Learnable Transformation
to address the time-varying outlier channels.
At the hardware level, 
we design the first low-bit FP quantizer and multiplier with lookup tables on FPGA and propose the first FPGA-based VAR accelerator featuring low-bit FP computation and an
elaborate two-level pipeline.
Extensive experiments show that compared to the state-of-the-art quantization method, our proposed FPQVAR
significantly improves Fréchet Inception Distance (FID) from 10.83 to 3.58, Inception Score (IS) from 175.9 to 
241.5 under 4-bit quantization.
FPQVAR also significantly improves the performance of 6-bit quantized VAR, bringing it on par with the FP16 model.
Our accelerator on AMD-Xilinx VCK190 FPGA
achieves a throughput of 1.1 image/s, which is $3.1\times$ higher than the integer-based accelerator.
It also demonstrates  $3.6\times$ and $2.8\times$ higher energy efficiency compared to the integer-based accelerator and GPU baseline, respectively.

\end{abstract}

\section{Introduction}
\label{sec:introduction}


Visual generation is an important task aiming to generate high-quality images or videos.
Recently, with the development of large language models (LLMs),
research on visual generation is shifting from diffusion models~\cite{ho2020denoising,song2020denoising, dhariwal2021diffusion, peebles2023scalable}
to autoregressive (AR) models~\cite{chen2020generative, esser2021taming, wu2024janus, chen2025janus} to leverage the scalability and generalizabilty of LLMs.
However, AR models suffer from inferior image generation quality and high latency, especially when generating high-resolution images, since they predict the next visual token in a raster-scan order.
To address this, visual autoregressive (VAR) modeling~\cite{tian2024visual}
has been proposed with next-scale prediction,
which generates the tokens of the next scale in a coarse-to-fine order.
Compared to diffusion models,
VAR has better performance and faster generation speed,
outperforming the powerful Diffusion Transformer (DiT)~\cite{peebles2023scalable} for the first time.
However, the large memory and computation cost of VAR
hinders its deployment on the resource-constrained devices.

\begin{figure}[!tb]
\centering
\includegraphics[width=\linewidth]{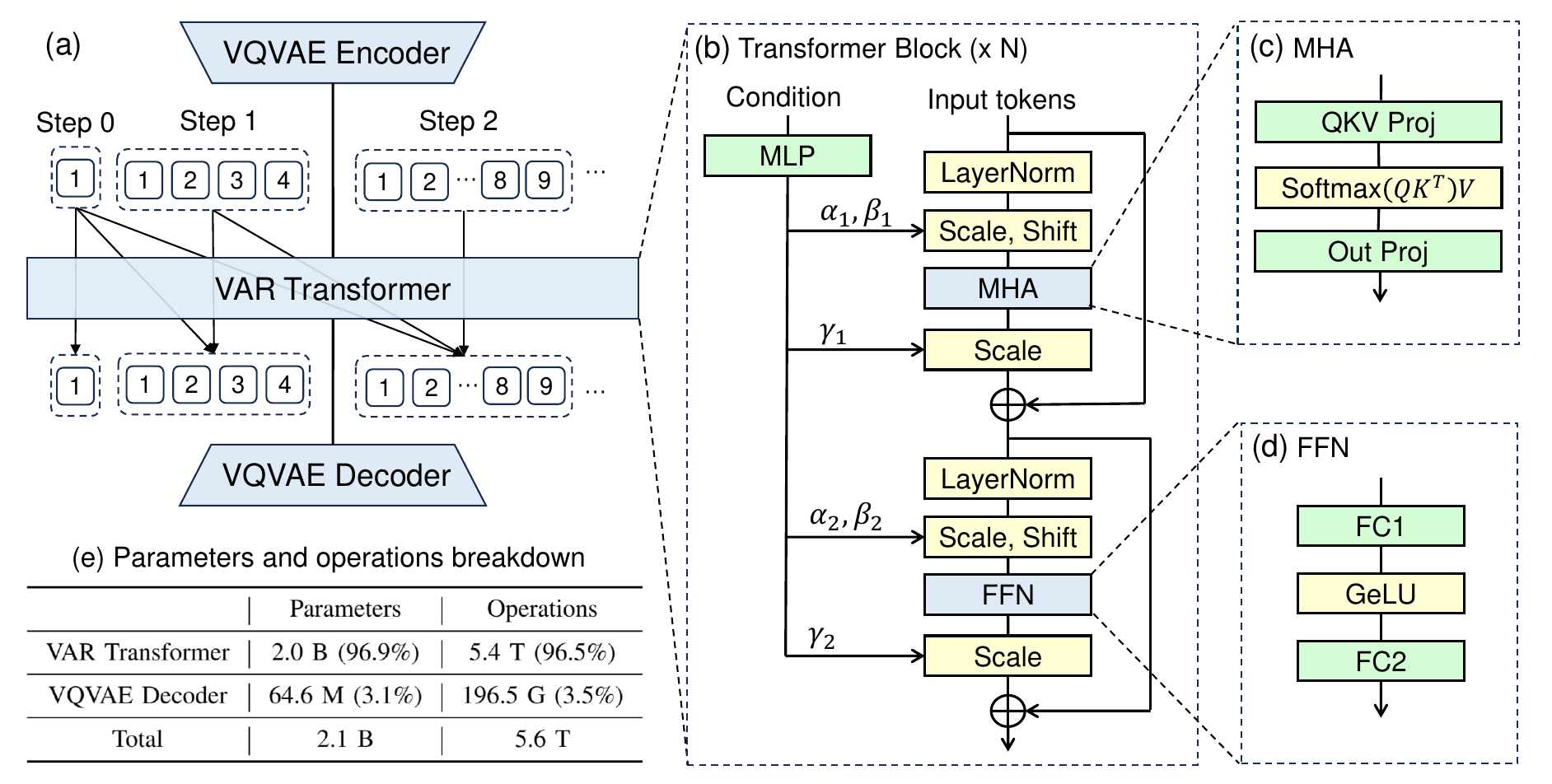}
\caption{VAR model architecture (a,b,c,d), its next-scale prediction (a), and the number of parameters and operations breakdown (e).} 
\label{fig:model_arch}
\vspace{-5pt}
\end{figure}

The model architecture of VAR is shown in Fig.~\ref{fig:model_arch}.
It consists of a VAR Transformer and a multi-scale VQVAE which also termed as tokenizer, including an encoder and a decoder.
During the image generation process,
the condition and initial token are sent to the VAR Transformer.
VAR Transformer autoregressively predicts a set of tokens from coarse to fine scales in ten steps.
At each step, all tokens of the current scale are predicted in parallel.
At last, tokens of the last step are sent to the VQVAE decoder to generate the image.

Although VAR reduces the number of image generation steps, the computation cost of each step is still heavy for the resource-constrained device. 
What's more, the memory consumption of VAR is also significant.
VAR-d30 model~\cite{tian2024visual} has more than 2 billion parameters that require more than 4GB memory in 16-bit floating point (FP16) format.
During inference time, 
it costs 5.6 TFLOPs to generate a single $256\times256$ image.
Such huge memory and computation requirements hinder its real-world applications.
Notably, as shown in Fig.~\ref{fig:model_arch}e, VAR Transformer constitutes majority of the total storage and computation consumption, accounting for approximately 97\%.
Therefore, reducing the cost of the VAR Transformer is our primary focus.

Quantization is a promising way to reduce both memory and computation cost 
by converting high-precision floating-point (FP) values to low-precision FP or integer (INT) representations.
It can be categorized into QAT (Quantization-Aware Training) and PTQ (Post-Training Quantization).
Since PTQ does not require re-training the entire model, it is more resource-efficient and has been widely studied for
LLMs~\cite{xiao2023smoothquant, wei2022outlier, ashkboos2024quarot, liu2024spinquant} and diffusion models~\cite{li2023q, so2023temporal,chen2024q,wu2024ptq4dit}.
However, for VAR, the research is still insufficient.
A recent study, LiteVAR~\cite{xie2024litevar}, discovers that FC2 layer in Fig.~\ref{fig:model_arch} is the quantization bottleneck and retains it in FP16 to improve the model performance.
However, this introduces a large amount of extra FP16 storage and computation, diminishing the advantages brought by quantization.
Moreover, LiteVAR suffers from significant performance degradation under low-bit quantization, e.g., 6-bit weight and activation (W6A6) and W4A4.

In this work, we identify the following challenges for quantizing VAR and propose our solutions to address them.
\textbf{\textit{1) Low-bit INT format incurs large quantization error for model quantization.}}
This is because the quantization levels of low-bit INT quantization are uniformly distributed, 
which is not suitable for the weights and activations in the neural network that normally exhibit Gaussian or Laplace distributions~\cite{kuzmin2022fp8,liu2023llm}.
To this end, we adopt the \textbf{low-bit FP quantization} for VAR.
\textbf{\textit{2) Highly imbalanced activation distribution of the FC2
layer incurs large quantization error.}}
We believe this is the key reason why LiteVAR~\cite{xie2024litevar} identifies the FC2 layer as the quantization bottleneck. 
To this end, we propose \textbf{Dual Format Quantization (DFQ)} for the imbalanced input activation of the FC2 layer,
which uses two different FP formats and scales for the negative and
positive values, respectively.
With imbalanced activation distributions being captured by DFQ, model performance is significantly improved, e.g., inception score (IS) increases from 167.0 to 189.5 and Fréchet Inception Distance (FID) improves from 10.72 to 8.25 for FP4 quantization.
However, there is still a performance gap between the low-bit FP model and the original FP16 model, which is attributed to the \textbf{\textit{3) time-varying outlier channels in the input activation of the QKV projection and FC1 layer}}.
Outliers stretch the quantization range and result in fewer effective quantization levels for other normal values, leading to large quantization error~\cite{xiao2023smoothquant}.
The temporal variation of outlier channels in VAR further increases the difficulty of quantization.
Thus, we propose \textbf{Group-wise Hadamard Transformation (GHT)} to reduce the time-varying outliers by applying Hadamard Transformation (HT)~\cite{ashkboos2024quarot} within each group.
Compared to HT, GHT significantly improves the hardware efficiency at the cost of performance loss.
To further enhance the model performance, we propose \textbf{GHT-Aware Learnable Transformation (GALT)}.
GALT effectively learns an optimal smoothing factor for all steps,
minimizing quantization error while ensuring no additional computational cost.
With GHT and GALT, model performance is significantly improved, e.g., IS of FP4 VAR increases from 189.5 to 241.5 and 
FID improves from 8.25 to 3.58.
Moreover, achieving efficient VAR inference on real device is important.
FPGA is a promising solution due to its efficiency and flexibility, 
which has been proven in many works~\cite{zeng2024flightllm, guo2024hg, liu2025flightvgm, li2025pushing}.
However, these works all focus on low-bit INT quantization. 
\textbf{\textit{4) To the best of our knowledge,  there has been no prior work exploring low-bit FP inference on FPGA.}}
To address this challenge, we propose the \textbf{efficient low-bit 
FP quantizer and multiplier design on FPGA} based on lookup tables (LUT).
Furthermore, we propose \textbf{the first FPGA-based VAR accelerator} co-designed with our proposed quantization algorithm.
Our FPQVAR accelerator features low-bit FP computation
and an elaborate two-level pipeline to reduce the overall latency.
As a result, our accelerator achieves a throughput of 1.1 image/s on 
AMD-Xilinx VCK190 FPGA,
which is $3.1\times$ faster than the integer-based accelerator.
It also demonstrates higher energy efficiency,
which is $3.6\times$ and $2.8\times$ compared to the integer-based accelerator and GPU baseline,
respectively.
Integrating the aforementioned methods, we present FPQVAR, an efficient post-training FP quantization framework for VAR featuring algorithm and hardware co-design. 
Our major contributions are summarized as follows:
\begin{itemize}
    \item We propose DFQ for the highly imbalanced input activation in VAR.
    We propose GHT and GALT to address the time-varying outlier channels, improving the model performance with minimal computational overhead.
    \item Our proposed quantization algorithm, FPQVAR, integrated with DFQ, GHT, and GALT, achieves the first plausible FP6 and FP4 VAR.
    Compared to the SOTA quantization method, FPQVAR improves FID from 10.83 to 3.58, IS from 175.9 to 
    241.5 under 4-bit quantization.
    FPQVAR also significantly improves the performance of 6-bit VAR, bringing it on par with FP16 model.
    \item We propose the first low-bit FP quantizer and multiplier on FPGA.
    We propose the first VAR accelerator featuring low-bit FP inference and an elaborate two-level pipeline, which achieves 1.1 image/s on AMD-Xilinx VCK190 FPGA and $2.8\times$ higher energy efficiency compared to the GPU baseline.

\end{itemize}

\section{related work and preliminary}
\label{sec:background}

\subsection{Visual Autoregressive Generation}
AR modeling~\cite{razavi2019generating, lee2022autoregressive,zheng2022movq} has been widely investigated as a promising alternative to diffusion models for image generation,
since the ability to harness the scalability and inherent language comprehension capability of LLMs.
AR models quantize image patches into discrete tokens leveraging VQVAE~\cite{van2017neural}
and use Transformer backbone, e.g., GPT to predict the next token.
However, they generate tokens in raster scan order,
which limits their performance and generation speed~\cite{tian2024visual}.
Recently, VAR~\cite{tian2024visual} has been proposed
with a paradigm shift in image generation from next-token prediction to next-scale prediction, leading to
better performance and faster generation speed
compared to previous AR methods.
VAR generates an image from coarse to fine scale, i.e, from resolution $(h_1, w_1)$ to $(h_K, w_K)$ in $K$ steps.
It outputs a set of tokens $r_k$ with resolution
$(h_k, w_k)$ at each step based on the previous sets of tokens
$(r_1, r_2, ..., r_{k-1})$ autoregressively as shown in Fig.~\ref{fig:model_arch}a.
The joint likelihood is formulated as:
\begin{equation}
 p(r_1, r_2, \dots, r_K) = \prod_{k=1}^{K} p(r_k \mid r_1, r_2, \dots, r_{k-1}).
\label{eq:joint_likelihood}
\end{equation} 
Following VAR, there are many works to further improve the visual generation quality.
For example, HART~\cite{tang2024hart} leverages continuous residual diffusion to compensate for the performance loss of the tokenizer in VAR.
Infinity~\cite{han2024infinity} replaces the vanilla tokenizer with an infinite-vocabulary bitwise tokenizer for better image quality.
However, these VAR-based models still face large memory and computation costs.
Quantization is an effective way to reduce the cost.
For VAR quantization, LiteVAR~\cite{xie2024litevar}
adopts the mixed-precision quantization scheme.
They keep the quantization-sensitive FC2 layer in FP16 to
improve the quantization performance.
However, it incurs large hardware cost and still has large performance degradation
under low-bit quantization, e.g., 6-bit and 4-bit.

\subsection{Floating-Point Quantization}

FP quantization has been a promising alternative to INT quantization, 
as it better aligns with the distribution of weights and activations in neural networks~\cite{kuzmin2022fp8}.
There are several works on low-bit FP quantization for vision models and LLMs.
~\cite{kuzmin2022fp8} first demonstrates FP8 is better than INT8 for neural network inference for small models like ResNet, ViT, and BERT.
~\cite{zhang2024integer} adopts a mixture of FP and INT formats quantization for LLMs.
~\cite{liu2023llm} quantizes LLM with FP4 format featuring optimal format for different layers and a pre-shifted exponent bias.
~\cite{zhang2023afpq} proposes to use two different scales
to quantize the asymmetric weights in LLM.
~\cite{wang2025optimizing} proposes an FP4 framework for LLM that achieves accuracy comparable to BF16.
However, these FP quantization methods cannot address the challenges for quantizing VAR
and there is also a lack of hardware implementation on FPGA.

In this work, we focus on quantizing the weights and activations of VAR with low-bit FP formats.
We first introduce the preliminaries of FP quantization here.
An FP number is composed of 1 sign bit (S), $j$ exponent bits (E), 
and $k$ mantissa bits (M), denoted as EjMk.
For FP formats with precision below 16 bits, IEEE 754 standard does not define the rules,
thus, we follow the rules in OCP Microscaling Formats (MX) Specification~\cite{rouhani2023microscaling}.
For an FP number, the encoding value is defined as follows:
\begin{equation}
v = 
\begin{cases}
(-1)^S \times 2^{E - bias} \times (1 + 2^{-k} \times M), & E > 0\\
(-1)^S \times 2^{1 - bias} \times (0 + 2^{-k} \times M), & E = 0
\end{cases}
\end{equation} 
where S, E, and M are the values of sign, exponent, and mantissa fields, respectively, 
$bias$ is the exponent bias, 
and $k$ is the number of mantissa bits.
OCP MX Specification
defines FP6-E2M3 ($bias=1$), FP6-E3M2 ($bias=3$), and FP4-E2M1.
Note that there are no encodings reserved for Inf or NaN in FP6 and FP4.
For FP4, 
we also follow~\cite{liu2023llm,wang2025optimizing} to take FP4-E1M2
and FP4-E3M0 into consideration.
Note that FP4-E1M2 here is equivalent to the INT quantization.
The encoding values of different FP4 formats are shown in Table~\ref{tab:FP4_values}.

\begin{table*}[!tb]
\centering
\caption{The encoding values of different FP4 formats.}
\label{tab:FP4_values}
\scalebox{1.0}{
\begin{tabular}{c|ccccccccccccccc}
\toprule
\multicolumn{1}{l}{} & \multicolumn{15}{c}{Binary Sequence}                                                                         \\ 
Format               & 1111 & 1110 & 1101 & 1100 & 1011 & 1010 & 1101  & 1000 / 0000 & 0001 & 0010 & 0011 & 0100 & 0101 & 0110 & 0111 \\ \midrule
E1M2                 & -3.5 & -3   & -2.5 & -2   & -1.5 & -1   & -0.5  & 0         & 0.5  & 1    & 1.5  & 2    & 2.5  & 3    & 3.5  \\ 
E2M1                 & -6   & -4   & -3   & -2   & -1.5 & -1   & -0.5  & 0         & 0.5  & 1    & 1.5  & 2    & 3    & 4    & 6    \\ 
E3M0                 & -16  & -8   & -4   & -2   & -1   & -0.5 & -0.25 & 0         & 0.25 & 0.5  & 1    & 2    & 4    & 8    & 16  \\ \bottomrule
\end{tabular}}
\end{table*}

To quantize the FP16 tensor $X$, 
we first calculate the scaling factor $s$ and map $X/s$
to the nearest FP encoding value following:
\begin{equation}
s = \frac{max(|X|)}{MAX_{fp}}, \quad X_{FP} = Q(\frac{X}{s}),
\end{equation}
where $MAX_{fp}$ is the maximum value in the FP format,
e.g. 6.0 for FP4-E2M1 or 28.0 for FP6-E3M2,
and $Q(\cdot)$ is the FP quantization function mapping values to the nearest grid.
The quantization granularity for INT quantization is also 
applicable for FP quantization.
In this work, 
we adopt per-channel weight quantization and per-token activation quantization for W8A8 and W6A6,
and per-group weight and activation quantization (group size=128) for W4A4,
which are the common configurations.

\section{Motivation}
\label{sec:motivation}

In this work,
we aim to quantize VAR to reduce the large memory and computation cost.
We identify the following four challenges through our analysis and 
preliminary experiments.

\begin{figure}[!tb]%
  \centering
  \hspace*{-3pt}
    \subfloat[]{
        \label{subfig:weight_distribution_and_quant_level}
        \includegraphics[width=0.24\textwidth]{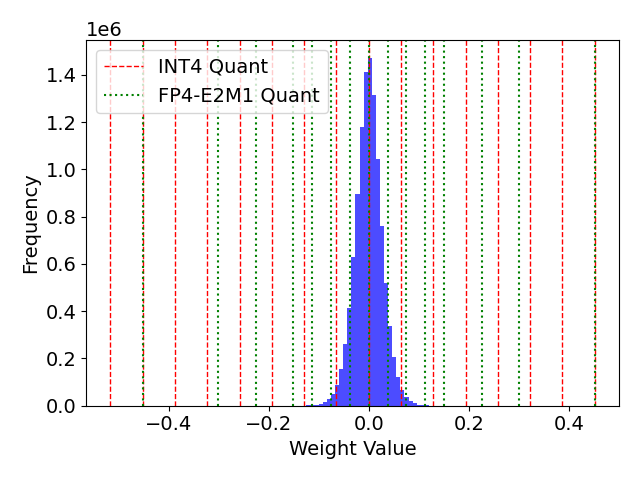}
      }
      \subfloat[]{
        \label{subfig:weight_quant_error}
        \includegraphics[width=0.23\textwidth]{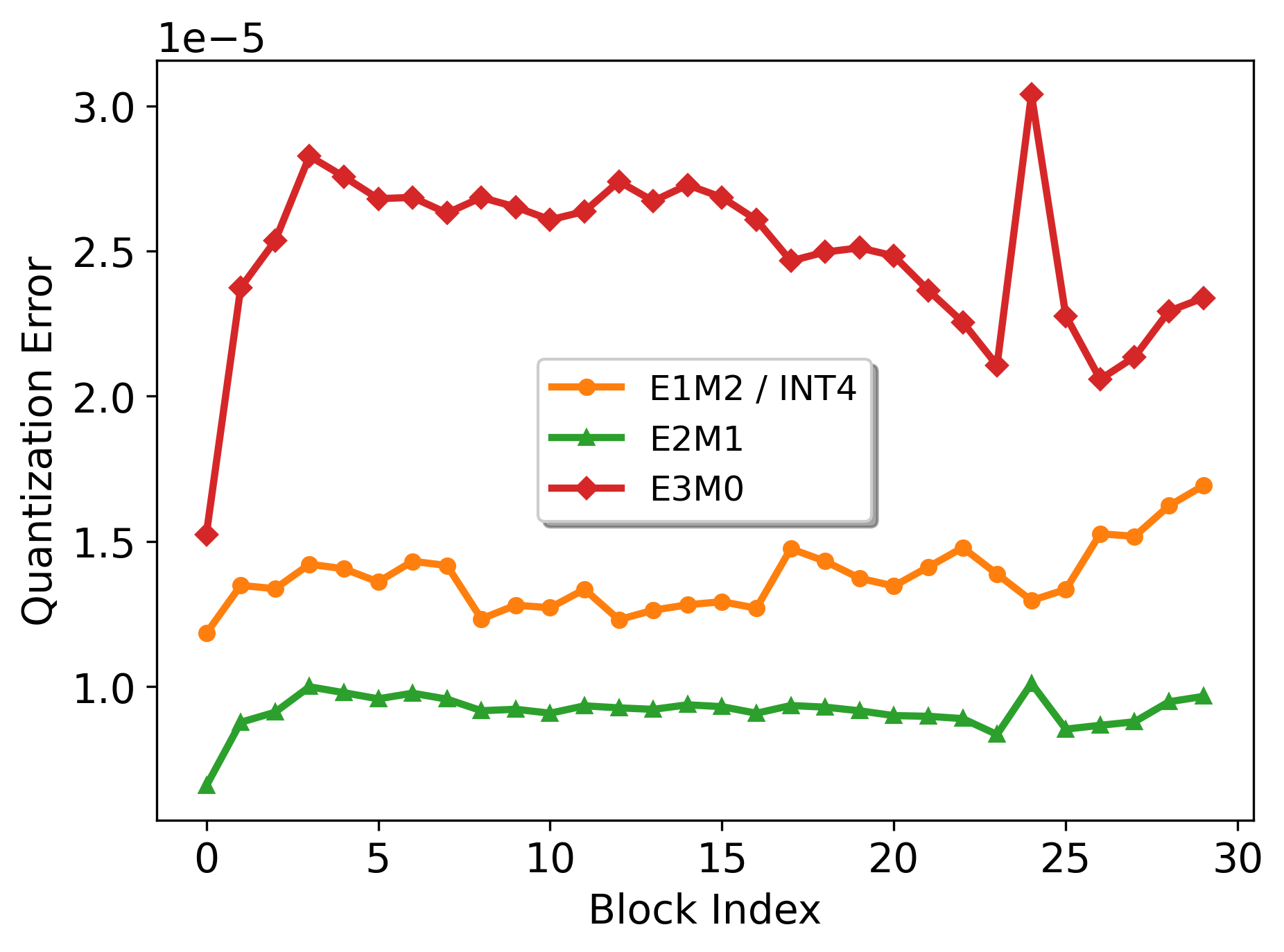}
      }
      
  \caption{
(a) Weight distribution of QKV projection layer in the 9-th block and the corresponding quantization levels for INT4 and FP4-E2M1.
(b) Quantization error of the QKV projection weight in each block with different quantization 
formats.
}
  \label{fig:observation 1}
   \vspace{-5pt} 
\end{figure}

\textbf{\textit{Challenge 1: Low-bit INT format incurs large quantization error for model quantization.}}
As shown in Fig.~\ref{subfig:weight_distribution_and_quant_level},
compared to INT quantization levels,
FP quantization levels are denser
near 0 and sparser as the values move away from 0, 
which better fits the Gaussian or Laplace distributions in neural networks.
Specifically,
FP4-E2M1 has $1.5\times$ lower quantization error compared to INT4 as 
shown in Fig.~\ref{subfig:weight_quant_error},
despite it has one fewer quantization level than INT4 (15 vs. 16).
Thus, we adopt low-bit FP formats instead of low-bit INT formats for VAR quantization.

We then conduct the preliminary experiments to verify
the effectiveness of FP quantization on model performance 
and build our FP baseline.
As shown in Table~\ref{tab:preliminary quantization},
RTN W8A8 quantization already achieves negligible accuracy degradation,
thus we focus on lower-bit quantization, e.g., 6-bit and 4-bit to achieve further memory saving and speedup.
For FP6 and FP4 quantization,
we set the quantization format the same for all layers to ensure hardware efficiency, i.e, to avoid the need for implementing different types of FP multiply-accumulate (MAC) units.
We find that for FP6 quantization, the optimal quantization format is E2M3 for weights and E3M2 for activations (W-E2M3/A-E3M2).
For FP4 quantization, the optimal quantization format is W-E2M1/A-E2M1.
Thus, we set them as our strong baseline.
Our FP6 W-E2M3/A-E3M2 baseline has already outperformed LiteVAR W6A6 significantly without bells and whistles.
Our FP4 W-E2M1/A-E2M1 baseline outperforms RTN W4A4 significantly by 102.4 in IS and 27.88 in FID,
which demonstrates the superiority of low-bit FP quantization.

\begin{table}[!tb]
\centering
\caption{Preliminary quantization results for VAR. RTN denotes round-to-nearest INT quantization. LiteVAR~\cite{xie2024litevar} adopts mixed precision INT quantization retaining FC2 layer in FP16.}
\label{tab:preliminary quantization}
\scalebox{0.9}{
\begin{tabular}{c|cccc}
\toprule
                         & IS $\uparrow$ & FID $\downarrow$   & Precision $\uparrow$ & Recall $\uparrow$ \\ \midrule
FP16                     & 302.0             & 1.98   & 0.82      & 0.60   \\ \midrule
RTN W8A8                 & 300.7             & 1.99   & 0.81      & 0.60   \\
RTN W6A6                 & 253.6             & 2.99   & 0.76      & 0.63   \\
RTN W4A4     & 64.6              & 38.60  & 0.53      & 0.58   \\ \midrule
LiteVAR W6A6             & 257.0             & 3.04   & 0.77      & 0.62   \\
LiteVAR W4A4             & 175.9             & 10.83  & 0.69      & 0.60   \\ \midrule
FP6 W-E2M3/A-E3M2    & 284.6             & 2.14   & 0.80      & 0.61   \\
FP4  W-E2M1/A-E2M1   & 167.0             & 10.72  & 0.69      & 0.60  \\ \bottomrule
\end{tabular}}
\end{table}

\textbf{\textit{Challenge 2: Highly imbalanced activation distribution of the FC2 layer incurs large quantization error.}}
LiteVAR~\cite{xie2024litevar} finds that quantizing FC2 layer leads
to large performance degradation,
thus retaining it in FP16.
Furthermore, we find that the difficulty in quantizing FC2 layer arises from the highly imbalanced activation distribution.
\begin{figure}[!tb]
\centering
\includegraphics[width=\linewidth]{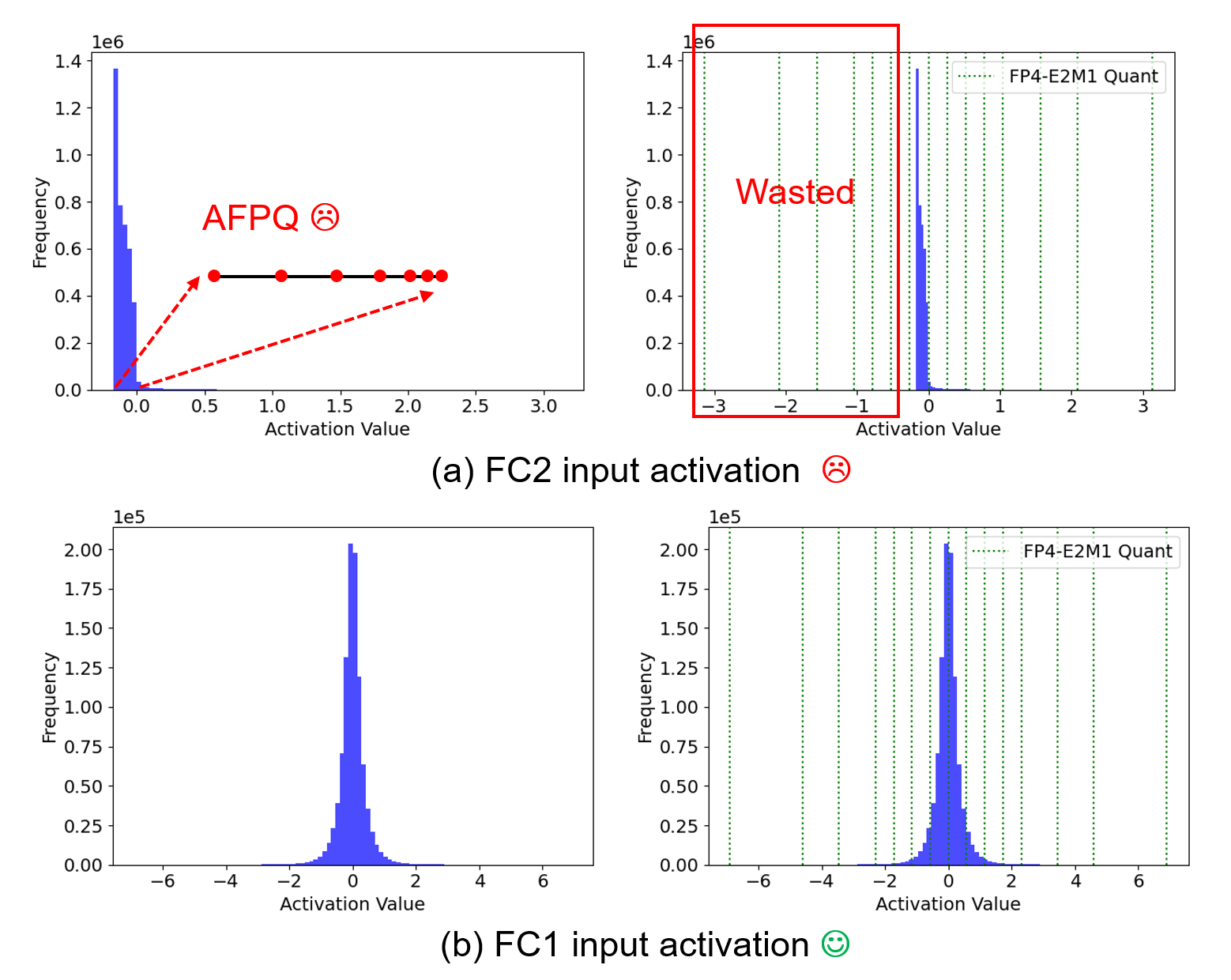}
\caption{Activation distributions of the FC1 and FC2 layer and the corresponding quantization levels.
Standard FP quantization and AFPQ~\cite{zhang2023afpq} can not fit in FC2 input activation distribution.} 
\label{fig:fc2_input_distribution}
\vspace{-5pt}
\end{figure}
As shown in Fig.~\ref{fig:fc2_input_distribution},
different from activations in other layers,
the input activation of FC2 layer is highly imbalanced,
which is because it is the output of the GeLU function as shown in Fig.~\ref{fig:model_arch}.
Specifically, there are 97.6\% of the values concentrated between -0.17 and 0, while only 2.4\% of the values are greater than 0 exhibiting a long-tail distribution.
Although standard FP quantization can adapt to the long-tail distribution in the positive field,
nearly half of the quantization levels are wasted in the negative field,
leading to large quantization error.
To address the asymmetry in FP quantization,
AFPQ~\cite{zhang2023afpq} proposes using two separate scales for the positive and negative part of weights, respectively.
However, it is suboptimal for the current situation
since the negative part of FC2 input activation peaks at -0.17 rather than 0.
While the quantization levels of AFPQ are still concentrated around 0 as shown in Fig.~\ref{fig:fc2_input_distribution}a, which would lead to inferior quantization performance.

\begin{figure*}[!tb]
\centering
\includegraphics[width=0.9\linewidth]{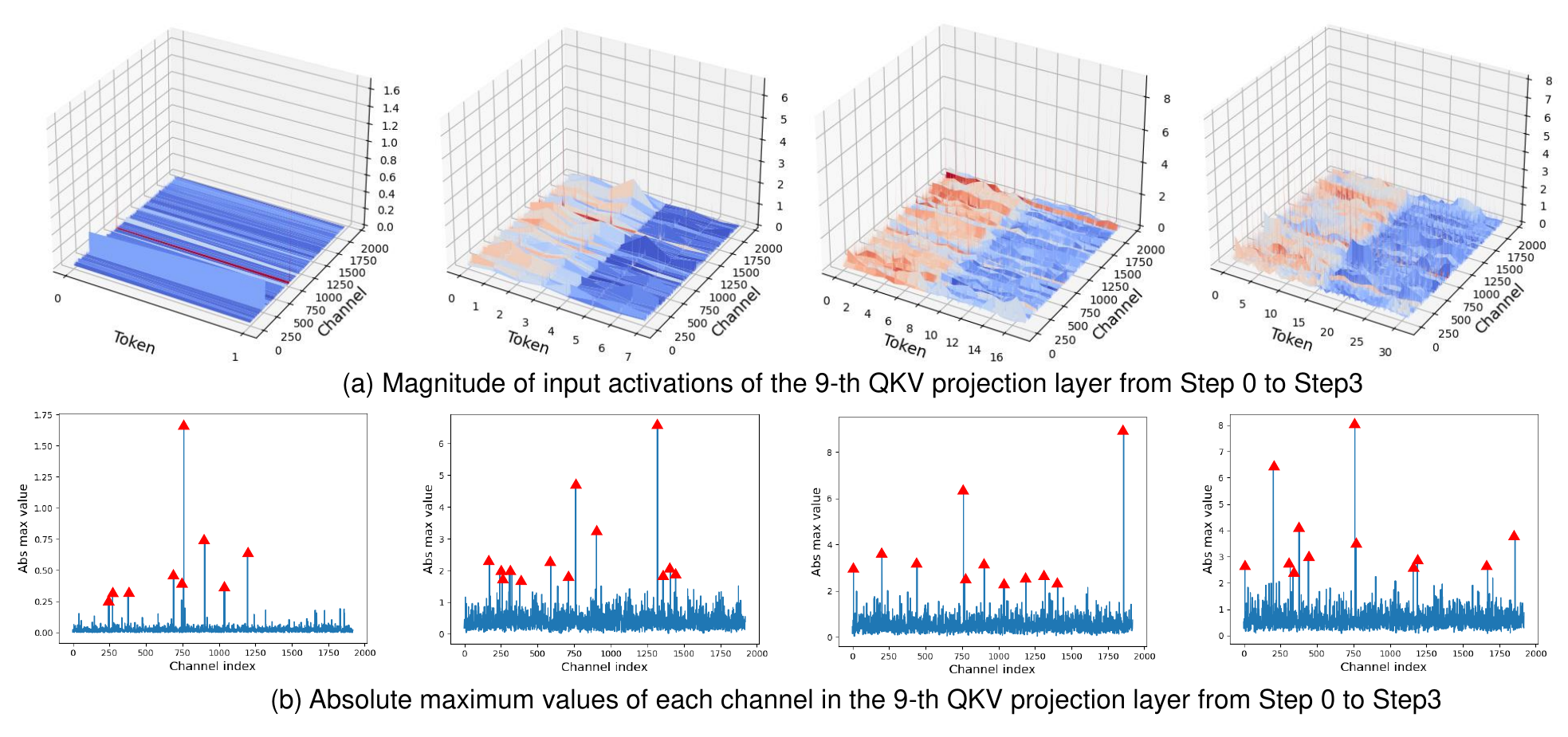}
\caption{Outlier channels (marked by red triangles) in VAR vary with the progression of time steps. Note that the number of tokens in the figure is doubled considering Classifier-Free Guidance (CFG)~\cite{ho2022classifier}. 
} 
\label{fig:outlier_channels}
\vspace{-5pt}
\end{figure*}

\textbf{\textit{Challenge 3: Time-varying outlier channels make quantization difficult.}}
We observe severe outliers in the input activations of the QKV projection and FC1 layer.
As shown in Fig.~\ref{fig:outlier_channels},
outliers persist in the fixed channels of activations in each step, 
which we termed as outlier channels.
These outliers incur large quantization error as in LLMs~\cite{xiao2023smoothquant,ashkboos2024quarot,liu2024spinquant}.
Moreover, the positions and magnitudes of outlier channels vary significantly across different steps 
as illustrated in Fig.~\ref{fig:outlier_channels}b.
Due to the temporal variation of outlier channels, smoothing-based quantization methods~\cite{xiao2023smoothquant,wei2022outlier,liu2023llm} become ineffective.
These methods use a smoothing factor $s$ to reduce the outliers in activation before quantization,
i.e., $Q(Xs^{-1})Q(sW)$.
However, they can only mitigate the impact of outliers for one time step and become ineffective for other time steps.
Applying a smoothing factor for each step is also infeasible.
Given that VAR involves 10 steps, 
this method needs to store $10\times$ quantized weight, resulting in the storage overhead even larger than the original FP16 weight.

\textbf{\textit{Challenge 4: Lack of efficient hardware implementation of low-bit FP computation units on FPGA.}}
FPGA is a promising candidate to achieve efficient inference for various models leveraging
quantization~\cite{zeng2024flightllm, yang2024sda, wei2025lightmamba, li2025pushing, guo2024hg}.
However, these works all focus on low-bit INT quantization.
There has been no prior work exploring low-bit FP
inference or VAR acceleration on FPGA, presenting these as unresolved challenges in the field.

To address the aforementioned challenges,
we propose FPQVAR,
an efficient post-training FP quantization framework
for VAR featuring algorithm and FPGA hardware co-design.
The proposed algorithm of FPQVAR includes: Dual Format Quantization (DFQ) for the highly imbalanced activation of FC2 layer in Sec.~\ref{subsec:DFQ},
Group-wise Hadamard Transformation (GHT) and GHT-Aware Learnable Transformation (GALT) to address the time-varying outlier channels in Sec.~\ref{subsec:Group-wise Hadamard Transformation} and~\ref{subsec:GHT-Aware Learnable Transformation}.
Hardware design of FPQVAR is in Sec.~\ref{subsec:Hardware Design}
including the first low-bit FP quantizer and multiplier on FPGA and the VAR accelerator featuring low-bit FP computation and an elaborate two-level pipeline.

\section{Method}
\label{sec:method}

\subsection{Dual Format Quantization}
\label{subsec:DFQ}

To quantize the highly imbalanced FC2 input activation,
we propose DFQ.
DFQ is based on the observation that the negative and positive parts of the activations exhibit distinct distributions as analyzed in Sec.~\ref{sec:motivation}.
Thus, we adopt two formats for the negative and positive parts, respectively,
capturing the different distributions.
DFQ is illustrated in Alg.~\ref{alg:DFQ},
which can be efficiently executed in parallel.
The formats of the negative and positive grids are searched
offline on the calibration dataset as shown in Alg.~\ref{alg:search optimal format}.
For example, the optimal FP4 DFQ format is the negative grids of FP4-E1M2 and the positive grids of FP4-E2M1.
Compared to AFPQ~\cite{zhang2023afpq}, our DFQ reduces the quantization error consistently as shown in Fig~\ref{subfig:Quantization error of FC2 activation}, achieving an average reduction of $1.7\times$.
It is worth noting that DFQ does not increase the complexity of the quantizer,
since the number of total quantization levels is the same as the standard FP quantization.

\begin{algorithm}[!tb]
\caption{Dual Format Quantization}
\label{alg:DFQ}
\begin{algorithmic}[1]
\State \textbf{Input}: $X$
\Comment{\textcolor{red}{FP16 activation}}
\State \textbf{Output}: $X_{q}$
\Comment{\textcolor{red}{Low-bit quantized activation}}
\State \textbf{Constant}: \textit{Neg\_grid, Pos\_grid} 

\State $X^- \gets \text{where}(X \le 0, X, 0)$, $X^+ \gets \text{where}(X > 0, X, 0)$
\State $s^- \gets \max(|X^-|) / \max(|Neg\_grid|)$ 
\State $s^+ \gets \max(|X^+|) / \max(|Pos\_grid|)$ 

\State $\hat{X}^- \gets X^- / s^-$, $\hat{X}^+ \gets X^+ / s^+$

\State $X_q^- \gets \text{quantize\_to\_nearest\_grid}(\hat{X}^-, Neg\_grid)$
\State $X_q^+ \gets \text{quantize\_to\_nearest\_grid}(\hat{X}^+, Pos\_grid)$

\State $X_{q} \gets X_q^- + X_q^+$
\State $X_{dequant} \gets X_q^-*s^- + X_q^+*s^+$ \Comment{\textcolor{red}{Dequantized tensor if needed}}

\State \Return $X_{q}$
\end{algorithmic}
\end{algorithm}

\begin{algorithm}[!tb]
\caption{Search for Optimal DFQ Format}
\label{alg:search optimal format}
\begin{algorithmic}[1]
\State \textbf{Input}: Calibration dataset $X$
\State \textbf{Output}: $\textit{optimal\_neg\_grid}$, $\textit{optimal\_pos\_grid}$
\State \textbf{Constant}: FP4 formats: $\{\text{E1M2}, \text{E2M1}, \text{E3M0}\}$

    \State Initialize: $\textit{min\_mse} \gets \infty$
    
    \For{$\textit{Neg\_grid} \in \{\text{FP4 formats}\}$}   
        \For{$\textit{Pos\_grid} \in \{\text{FP4 formats}\}$}
            \State $X_{dequant} \gets \text{DFQ}(X, \textit{Neg\_grid}, \textit{Pos\_grid})$
            \State $\textit{current\_mse} \gets \text{MSE}(X_{dequant}, X)$
            
            \If{$\textit{current\_mse} < \textit{min\_mse}$}
                \State $\textit{min\_mse} \gets \textit{current\_mse}$
                \State $\textit{optimal\_neg\_grid} \gets \textit{Neg\_grid}$
                \State $\textit{optimal\_pos\_grid} \gets \textit{Pos\_grid}$
            \EndIf
        \EndFor
    \EndFor
    
    \State \Return ($\textit{optimal\_neg\_grid}$, $\textit{optimal\_pos\_grid}$)
\end{algorithmic}
\end{algorithm}


\begin{figure}[!tb]%
  \centering
    \subfloat[Quantization error of FC2]{
        \label{subfig:Quantization error of FC2 activation}
        \includegraphics[width=0.22\textwidth]{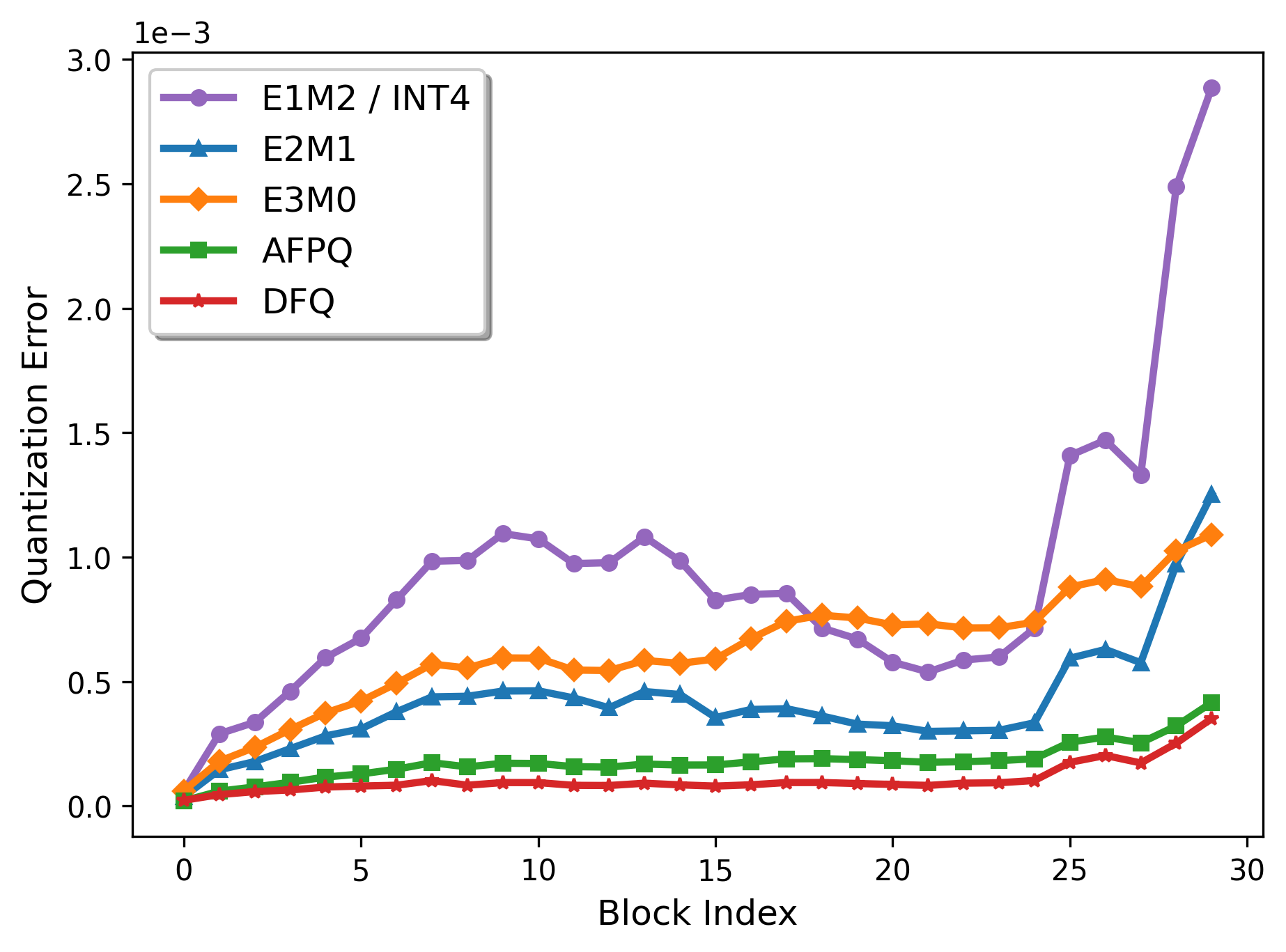}
      }
      \subfloat[Calibration dataset construction]{
        \label{subfig:Calibration dataset construction}
        \includegraphics[width=0.24\textwidth]{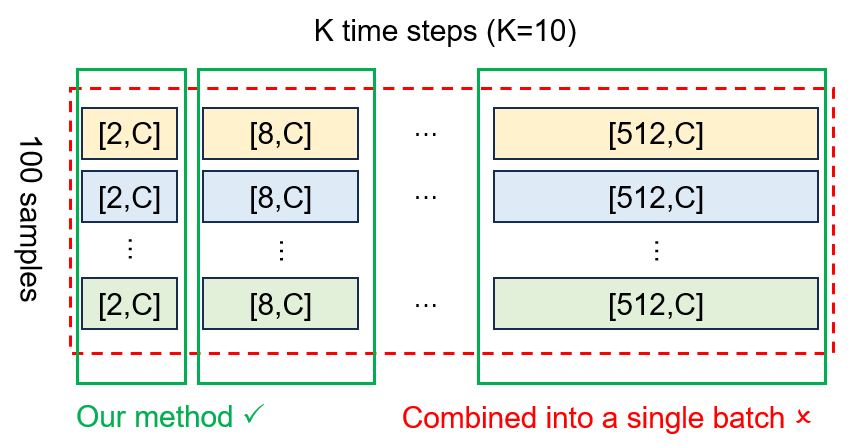}
      }
      
  \caption{
(a) Quantization error of FC2 input activation with different quantization formats across all blocks.
(b) Our proposed method for constructing the calibration dataset in Sec.~\ref{subsec:GHT-Aware Learnable Transformation},
only one layer is plotted here for clarity.
}
  \label{fig:rotate activation}
  \vspace{-5pt}
\end{figure}

\subsection{Group-wise Hadamard Transformation} 
\label{subsec:Group-wise Hadamard Transformation}
In VAR, we observe time-varying outlier channels in
QKV projection and FC1 layer,
which incurs large quantization error and inferior model performance.
A straightforward approach is to apply the rotation-based method~\cite{ashkboos2024quarot} on VAR.
The computation of QKV projection or FC1 layer before and after applying the rotation, i.e., Hadamard Transformation (HT) is as follows:
\begin{equation}
\label{eq:hadamard rotation}
\scalebox{0.85}{$\displaystyle
[\frac{X}{\|X\|}\odot(1+\alpha)+\beta]W^T  \Rightarrow Q([\frac{X}{\|X\|}\odot(1+\alpha)+\beta] H) Q({H^T} W^T),
$}
\end{equation}
where $X$ is the input activation to the LayerNorm before the linear layer, $W$ is the weight of linear layer, $Q(\cdot)$ is the quantization function, $\alpha$ and 
$\beta$ are the adaptive scale and shift factor, respectively.
The rotation-based method multiplies the activation and weight of the linear layer with Hadamard matrix $H$ and its transpose $H^T$, respectively,
to reduce outliers while maintaining the computational invariance.
Since $H$ is independent of the specific time step, this method is effective in reducing time-varying outliers.
After rotation, outliers are significantly reduced 
comparing Fig.~\ref{fig:outlier_channels}a and Fig.~\ref{fig:rotate activation}a.

\begin{figure}[!tb]%
  \centering
  \hspace*{-3pt}
    \subfloat[After HT]{
        \label{subfig:After rotation}
        \includegraphics[width=0.24\textwidth]{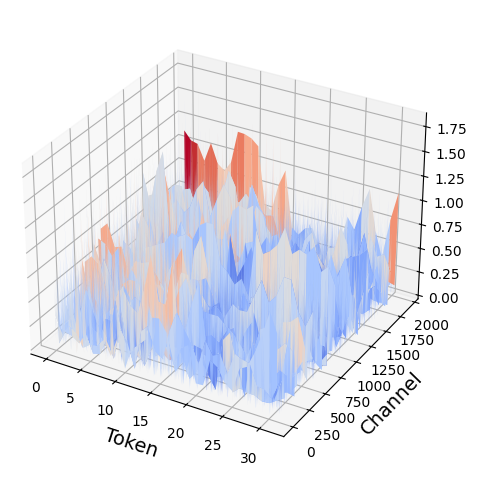}
      }
  \hspace*{-5pt}
      \subfloat[After GHT]{
        \label{subfig:After block-diagonal rotation}
        \includegraphics[width=0.24\textwidth]{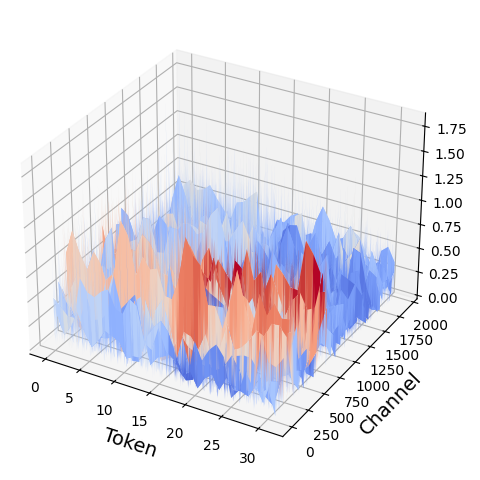}
      }
      
  \caption{
Activation distribution of the 9-th QKV projection layer at Step 3 after (a) HT and (b) our proposed GHT.
}
  \label{fig:rotate activation}
  \vspace{-5pt}
  
\end{figure}

However, the naive rotation method is hardware inefficient.
Although $H^T$ can be fused into $W$ offline,
$H$ is forced to be executed online for VAR,
which adds a lot of computational overhead.
For the rotation method in LLM~\cite{ashkboos2024quarot}, $H$ can be fused into $X$ on the condition that $\alpha$ can be absored into $W^T$, which is
$Q([\frac{X{H}}{\|X\|}] ) Q({H^T} diag(\alpha) W^T)$.
However, unlike LLM,
the scale factor $\alpha$ in VAR is generated through the MLP layer based on the input condition
at runtime,
preventing $(1+\alpha)$ from being absored into $W$ in advance.
Thus, $H$ is enforced to be multiplied online to the input of QKV projection and FC1 layer.

In order to reduce the computational overhead,
we propose Group-wise Hadamard Transformation (GHT).
Specifically, we replace the Hadamard matrix $H \in \mathbb{R}^{C \times C}$ in Eq.~\ref{eq:hadamard rotation}, where $C$ is the model dimension,
with a block-diagonal Hadamard matrix $H_B = \text{BlockDiag}(H_{b1}, H_{b2}, \dots, H_{bn})$,
where $H_{bi} \in \mathbb{R}^{128 \times 128}, n=C/128, i=1,2...,n$.
We set $H_{b1}$ to $H_{bn}$ as the same ${128 \times 128}$ Hadamard matrix for better hardware efficiency.
The proposed rotation process can be viewed
as dividing the activation $X \in \mathbb{R}^{T \times C}$ into multiple groups of size $T\times 128$,
and performing HT within each group.
When $C$ equals 1920, GHT reduces the FLOPs of HT by $15\times$.
Moreover, since the group size of GHT aligns with the group size of quantization, 
the rotation, quantization, and matrix multiplication (MM) can be executed in a pipeline as shown in Fig.~\ref{fig:pipeline}c.
Compared to the HT implementation in~\cite{wei2025lightmamba}, our GHT reduces the latency by $3.9\times$ as shown in Fig.~\ref{fig:pipeline}b and Fig.~\ref{fig:pipeline}c.

\subsection{GHT-Aware Learnable Transformation}
\label{subsec:GHT-Aware Learnable Transformation}

Although GHT is more hardware friendly,
it is less effective in reducing outliers compared to HT,
thus leading to inferior model performance.
As shown in Fig.~\ref{fig:rotate activation},
compared to the activation tensor after HT,
the activation tensor after GHT contains 
more elements with large values indicated by red color, 
which will affect the quantization of a larger number of normal values.
The reason is that GHT only amortizes the outliers within each group
instead of whole input channels.
To further reduce the quantization error and improve the model performance,
we propose GHT-Aware Learnable Transformation (GALT).
Our key idea is to learn a smoothing factor $\lambda \in \mathbb{R}^{1 \times C}$ for each QKV projection and FC1 layer that minimizes the output quantization error for all steps.
We use a gradient-based optimization method.
The optimization objective is defined as:
\begin{equation}
\label{eq:learnable transformation MSE}
\scalebox{0.9}{$\displaystyle
\underset{\lambda}{\arg\min} \sum_{i = 0}^{K - 1}  \left\| X_i W^T - Q(X_i \lambda H_B) Q(H_B^T \lambda^{-1} W^T) \right\|_2,
$}
\end{equation}

where $K$ is the total number of steps, 
$X_i$ is the input activation at Step $i$,
$W$ is the weight,
$H_B$ is the block-diagonal Hadamard matrix,
$Q(\cdot)$ is the FP quantization function.
Since $Q(\cdot)$ is non-differentiable,
we leverage Straight-Trough Estimator (STE)~\cite{bengio2013estimating}
to bypass the gradient during back-propagation in our optimization.

We perform the optimization on a calibration dataset.
We find that constructing the calibration dataset for VAR is challenging, 
which is crucial for the effectiveness and efficiency of the optimization process.
As shown in Fig.~\ref{subfig:Calibration dataset construction},
the input activations of each layer have different dimensions across different time steps, 
thus they cannot be combined into a single batch.
On the other hand, individually treating each activation as a calibration data like that for LLMs would result in an excessively slow optimization process since there are multiple time steps.
Thus, we propose our construction method shown in Fig.~\ref{subfig:Calibration dataset construction}.
It is illustrated below, considering one layer for clarity.
We choose 100 samples randomly.
For each sample, we run the forward pass and collect its input activations across all K time steps,
resulting in K activations.
Then, for each time step, we concatenate these activations from different samples.
Consequently, we build our calibration dataset, 
which for each layer consists of K calibration data, each with a dimension of $[100, T_i, C]$, where $i \in \{0, 1, \dots, K-1\}$.

The optimization process of GALT is illustrated in Alg.~\ref{alg: GALT optimize}.
Note that we do not directly use Eq.~\ref{eq:learnable transformation MSE} as the loss function, but use MSE at each time step as the loss function and update over K iterations per epoch.
This strategy enhances convergence speed and results in a smaller final loss.
With GALT, model performance is significantly improved.
Moreover, GALT does not incur extra computational overhead during inference.
Specifically, $\lambda^{-1}$ can be fused into $W$ offline
and $\lambda$ can be absorbed into the weights and bias of the MLP layer in Fig.~\ref{fig:model_arch} as:
\begin{equation}
\label{eq:fuse GALT}
\scalebox{0.9}{$\displaystyle
\begin{split}
X_i\lambda = [\frac{X'_i}{\|X'_i\|}\odot(1+\alpha)+\beta]\lambda &= [\frac{X'_i}{\|X'_i\|}\odot(\lambda+\alpha\lambda)+\beta\lambda] \\ 
&= [\frac{X'_i}{\|X'_i\|}\odot(\lambda+\hat{\alpha})+\hat{\beta}],
\end{split}
$}
\end{equation}
where $X_i$ is the input of the linear layer, $X'_i$ is the input of the preceding LayerNorm, $\hat{\alpha}$ and $\hat{\beta}$ are the new scale and shift factor generated from the MLP layer with $\lambda$  fused.

\begin{algorithm}[!tb]
\caption{Optimization of GALT}
\label{alg: GALT optimize}
\begin{algorithmic}[1]
\State \textbf{Input}: Calibration dataset $D$, $W$, $H_B$ 
\State \textbf{Output}: $L$ \Comment{ \textcolor{red}{A list of  $best\_\lambda$ for each layer}}
\For{$l = 1$ \textbf{to} \text{$number\_of\_layers$}}
    \State Initialize: $best\_loss \gets \infty$, $\lambda \gets \mathbf{1}_{1 \times C}$
    \For{$i = 1$ \textbf{to} $epochs$}
        \State $epoch\_loss \gets 0.0$
        \For{$j = 0$ \textbf{to} $K-1$}
            \State $loss \gets \text{MSE}(D[l][j], W[l], \lambda, H_B)$
            \Comment{\textcolor{red} {Based on Eq.~\ref{eq:learnable transformation MSE}}}
            \State Gradient update
            \State $epoch\_loss \gets epoch\_loss + loss$
        \EndFor
        \If{$epoch\_loss < best\_loss$}
            \State $best\_loss \gets epoch\_loss$, $best\_\lambda \gets \lambda$
        \EndIf
    \EndFor
    \State $L.$append$(best\_\lambda)$
\EndFor
\State \Return $L$
\end{algorithmic}
\end{algorithm}

\subsection{Hardware Design}
\label{subsec:Hardware Design}

To achieve efficient low-bit FP inference of VAR on FPGA,
we propose our hardware design co-designed
with our algorithms.

\begin{figure*}[!tb]
\centering
\includegraphics[width=0.95\linewidth]{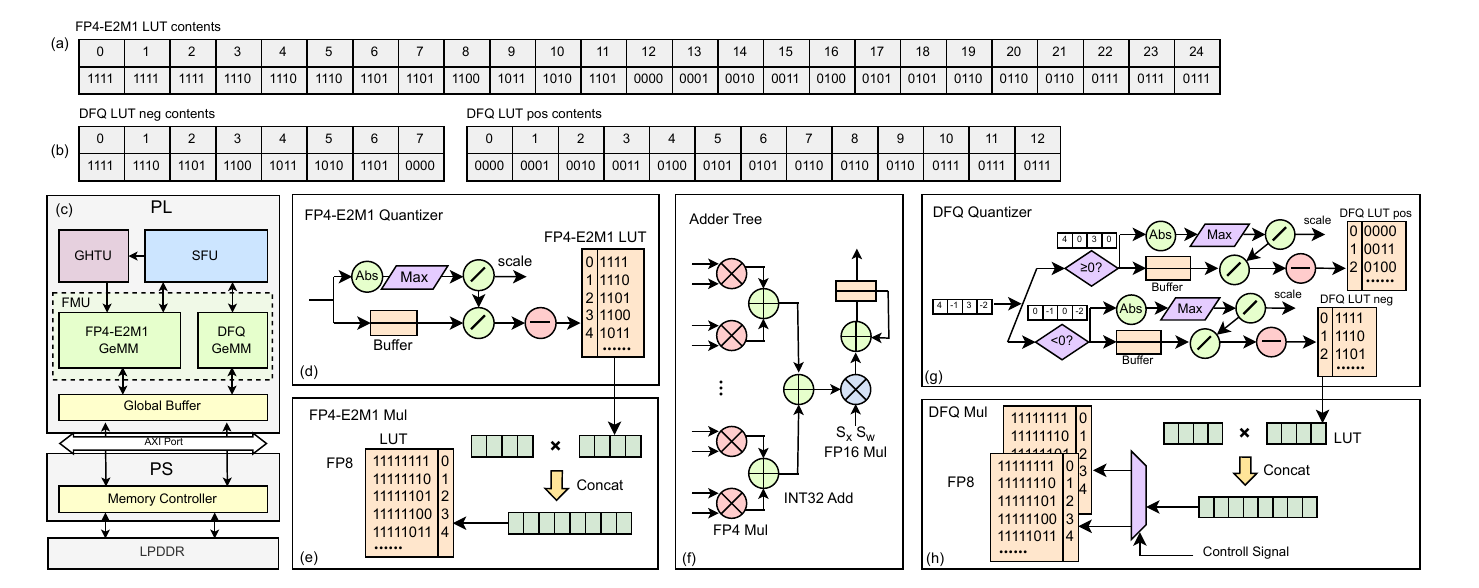}
\caption{Hardware architecture of our VAR accelerator. 
} 
\label{fig:hardware}
\vspace{-5pt}
\end{figure*}

\textbf{FP Quantizer and Multiplier}
First, as there is no existing low-bit FP computation units for FPGA, 
we propose the FP4 quantizer including the vanilla FP4-E2M1 quantizer and DFQ quantizer. 
As shown in Fig.~\ref{fig:hardware}d, 
an FP4-E2M1 quantizer involves two passes.
The top pass calculates the scaling factor.
The bottom pass scales the input tensor to $[-12, 12]$ with the scaling factor,
then subtracts it by -12, 
and uses 4 LUT5s (LUT5: 5-bit input, 1-bit output) to generate the 4-bit binary encodings.
The LUT contents are shown in Fig.~\ref{fig:hardware}a.
It is constructed by 
multiplying the E2M1 encoding values in Table~\ref{tab:FP4_values} by 2,
which intends to transform the non-integer into an integer, e.g., 0.5 to 1,
and then combining with the corresponding binary encodings.
The proposed DFQ quantizer is shown in Fig.~\ref{fig:hardware}g.
Based on our Alg.~\ref{alg:DFQ},
the positive and negative values are scaled separately
and look up their binary encodings through the DFQ\_LUT\_pos and DFQ\_LUT\_neg, respectively,
each consumes 4 LUT4s.
The FP multiplier is also built with LUTs.
Specifically, we concatenate the two 4-bit inputs together to form an address, which is then used to query the multiplication results in LUT, which are in FP8 format.
Then we convert FP8 to INT32 and perform accumulation.
After that, we multiply the results with the scaling factors of activations and weights
as shown in Fig.~\ref{fig:hardware}f.

\begin{figure}[!tb]
\centering
\includegraphics[width=1.0\linewidth]{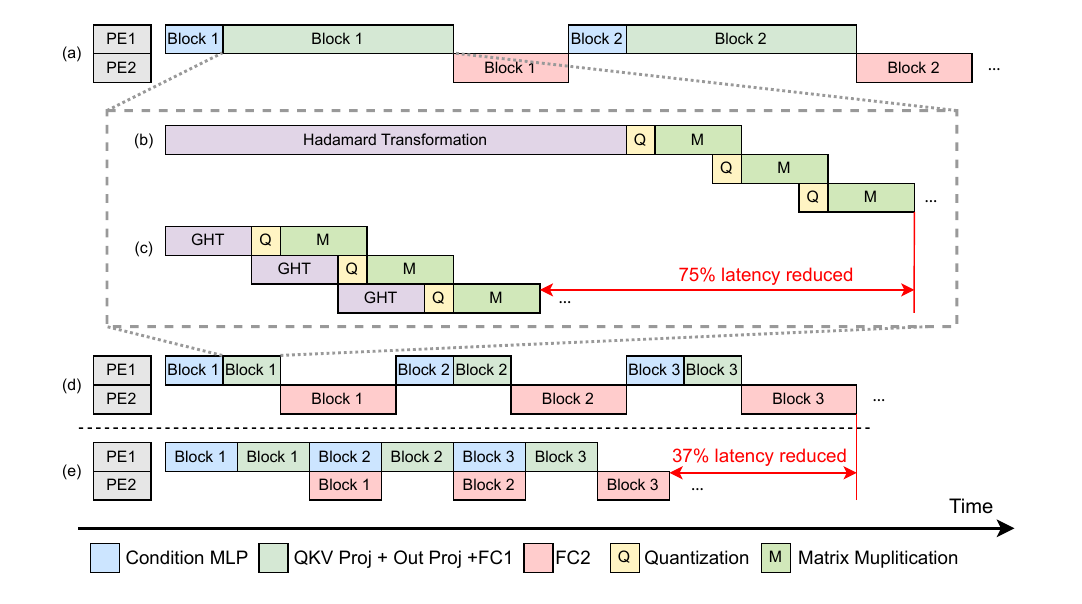}
\caption{The proposed two-level pipeline.} 
\label{fig:pipeline}
\vspace{-5pt}
\end{figure}

\textbf{VAR Accelerator} The architecture of our proposed VAR accelerator is shown in Fig.~\ref{fig:hardware}c.
It mainly consists of three components: FP Matrix Multiplication Unit (FMU), Group-wise Hadamard Transformation Unit (GHTU), and Special Function Unit (SFU).
We unfold one block of VAR on FPGA.
SFU handles the computation of nonlinear operators, such as LayerNorm, Softmax, and GeLU,
inspired by the design in~\cite{li2025pushing,guo2024hg}. 
GHTU performs group-wise Hadamard transformation leveraging a 128-point Fast Hadamard Transformation unit,
inspired by~\cite{wei2025lightmamba}.
FMU contains two General Matrix Multiplication (GeMM) units, i.e, DFQ GeMM unit for FC2 layer, and 
FP4-E2M1 GeMM unit for other linear layers.
We also propose a two-level pipeline to reduce the overall latency.
The first-level pipeline is shown in Fig.~\ref{fig:pipeline}b and Fig.~\ref{fig:pipeline}c, enabled by GHT.
Since GHT does not rely on the whole input tensor like HT,
we can perform rotation, quantization, and MM group-by-group in pipeline.
We propose the second-level pipeline shown in Fig.~\ref{fig:pipeline}d and Fig.~\ref{fig:pipeline}e.
Since the Condition MLP in VAR is only dependent on the input condition, the computation of Condition MLP of the $(i+1)$-th Block with FP4-E2M1 GeMM Unit can be overlapped with the computation of FC2 layer of the $i$-th Block with DFQ GeMM Unit with careful parallelism design.
By combining these two-level pipelines,
we reduce the total latency by $2.3\times$ from Fig.~\ref{fig:pipeline}a to Fig.~\ref{fig:pipeline}e.

\section{experiment}
\label{sec:experiment}

\subsection{Experimental Setup}
\label{subsec:Experimental Setup}

\textbf{Algorithm}
We evaluate our quantization algorithm under W6A6 and W4A4 settings on ImageNet $256\times256$ and $512\times512$ conditional generation benchmark.
We employ four evaluation metrics: Inception Score (IS)~\cite{salimans2016improved}, 
Fréchet Inception Distance (FID)~\cite{heusel2017gans}, Precision, and Recall over 50,000 images for each method.
For our GALT,
we construct a calibration dataset following the method in Sec.~\ref{subsec:GHT-Aware Learnable Transformation}.
We utilize the AdamW~\cite{loshchilov2017decoupled} optimizer with zero weight decay.
The number of epochs is 50.
The learning rate is 0.01 without learning rate decay.

\textbf{Hardware}
We implement our hardware design on the AMD-Xilinx VCK190 FPGA without leveraging the AI engine.
We use Vitis HLS and Vivado design flow.
We measure the throughput with PYNQ framework,
and power consumption with Xilinx’s BEAM tool.

\subsection{Experimental Results}
\label{subsec:Experimental Results}

\begin{table}[!tb]
\centering
\caption{Quantitative comparison between different quantization methods 
on ImageNet $256\times256$ benchmark.}
\label{tab:algorithm result}
\scalebox{0.95}{
\begin{tabular}{c|cccc}
\toprule
                         & IS $\uparrow$ & FID $\downarrow$   & Precision $\uparrow$ & Recall $\uparrow$ \\ \midrule
FP16              & 302.0          & 1.98          & 0.82          & 0.60          \\ \midrule
INT6 RTN          & 253.6          & 2.99          & 0.76          & \textbf{0.63}         \\
INT6 LiteVAR      & 257.0          & 3.04          & 0.77          & 0.62          \\
FP6 baseline      & 284.6          & 2.14          & 0.80          & 0.61          \\
FP6 MixedFormat   & 289.4          & 2.09          & 0.80          & 0.60          \\
\rowcolor{gray!20}
FP6 FPQVAR (ours) & 295.0  & 2.03      & 0.80          & 0.61           \\ 
\rowcolor{gray!20}
FP6 FPQVAR KV6 (ours) & \textbf{296.8}    & \textbf{2.00}      & \textbf{0.81}          & 0.61          \\
\midrule
INT4 RTN          & 64.6           & 38.60         & 0.53          & 0.58          \\
INT4 LiteVAR      & 175.9          & 10.83         & 0.69          & 0.60          \\
FP4 baseline      & 167.0          & 10.72         & 0.69          & 0.60          \\
FP4 MixedFormat   & 177.1          & 8.80          & 0.70           & 0.60          \\
\rowcolor{gray!20}
FP4 FPQVAR (ours) & \textbf{241.5} & \textbf{3.58} & \textbf{0.77} & \textbf{0.60} \\
\rowcolor{gray!20}
FP4 FPQVAR KV6 (ours) & 241.3      & 3.64          & \textbf{0.77}         & \textbf{0.60}         \\  
\bottomrule
\end{tabular}}
\end{table}

\begin{table}[!tb]
\centering
\caption{Quantitative comparison between different quantization methods 
on ImageNet $512\times512$ benchmark.}
\label{tab:algorithm result 512x512}
\scalebox{0.95}{
\begin{tabular}{c|cccc}
\toprule
                         & IS $\uparrow$ & FID $\downarrow$   & Precision $\uparrow$ & Recall $\uparrow$ \\ \midrule
FP16              & 291.9          & 2.72         & 0.82          & 0.56          \\ \midrule
INT6 RTN          & 161.3          & 11.77          & 0.74          & \textbf{0.63}         \\
INT6 LiteVAR      & 179.5          & 9.47          & 0.76          & 0.60          \\
FP6 baseline      & 272.7          & 3.37         & 0.81          & 0.56          \\
\rowcolor{gray!20}
FP6 FPQVAR (ours) & 287.7           & \textbf{2.83}      & \textbf{0.81}          & 0.54           \\ 
\rowcolor{gray!20}
FP6 FPQVAR KV6 (ours) & \textbf{289.5}    & \textbf{2.83}      & \textbf{0.81}          & 0.55          \\
\midrule
INT4 RTN          & 78.5           & 35.26         & 0.61          & \textbf{0.61}          \\
INT4 LiteVAR      & 148.4          & 16.56        & 0.73          & 0.59          \\
FP4 baseline      & 95.8          & 30.30         & 0.64          & 0.59          \\
\rowcolor{gray!20}
FP4 FPQVAR (ours) & 184.3           & 9.24      & \textbf{0.75}        & 0.57 \\
\rowcolor{gray!20}
FP4 FPQVAR KV6 (ours) & \textbf{186.6}      & \textbf{9.21}          & \textbf{0.75}         & 0.58         \\  
\bottomrule
\end{tabular}}
\end{table}

\textbf{Algorithm Evaluation}
We compare our method FPQVAR with the SOTA quantization method for VAR, LiteVAR~\cite{xie2024litevar}, which adopts mixed-precision quantization technique retaining FC2 layer in FP16.
We also compare with our FP baseline built in Sec.~\ref{sec:motivation} 
and FP mixed-format quantization,
which searches for the optimal FP formats for each layer to minimize the quantization error as in~\cite{liu2023llm}.
All methods are evaluated under the same random seed setting on the same GPU machine.
Note that the FP16 result is evaluated with the pretrained FP16 VAR-d30 model,
however, it slightly differs from the results presented in the original paper~\cite{xie2024litevar}
due to different random seed values on different GPU machines.
The results of the $256\times256$ benchmark are shown in Table~\ref{tab:algorithm result}.
In the W6A6 setting, 
compared to the SOTA method LiteVAR,
our proposed FPQVAR improves IS by 38.0 and FID by 1.01.
What's more, the performance of FP6 FPQVAR is on par with the FP16 model. 
In the W4A4 setting,
our FPQVAR significantly outperforms LiteVAR,
improving IS by 65.6 and FID by 7.25.
FPQVAR also surpasses the strong FP mixed-format quantization method in both W6A6 and W4A4 settings.
Moreover, we quantize the key and value cache (KV Cache) to FP6-E2M3 to further reduce the memory cost.
Results show that our KV Cache quantization incurs negligible performance degradation,
or even better performance for 6-bit quantization.
Our method also demonstrates consistent superiority on the $512\times512$ benchmark.
As shown in Table~\ref{tab:algorithm result 512x512},
in the W6A6 setting,
our FPQVAR performs on par with the FP16 model
and outperforms the SOTA method LiteVAR by 110.2 in IS and 6.64 in FID.
In the W4A4 setting,
compared to LiteVAR,
our method improves IS by 35.9 and FID by 7.32.
With KV Cache quantized to FP6,
the model performance is even slightly improved.
Note that we only apply GALT for FP4 FPQVAR and do not apply it for FP6 FPQVAR,
since we find that FP6 quantization is already capable of quantizing the activations after GHT,
therefore employing GALT on it has little impact.
Another interesting finding is that compared to the FP16 model,
quantization enhances VAR’s recall, 
which suggests it generates more diverse realistic images.

\begin{figure*}[!tb]
\centering
\includegraphics[width=0.92\linewidth]{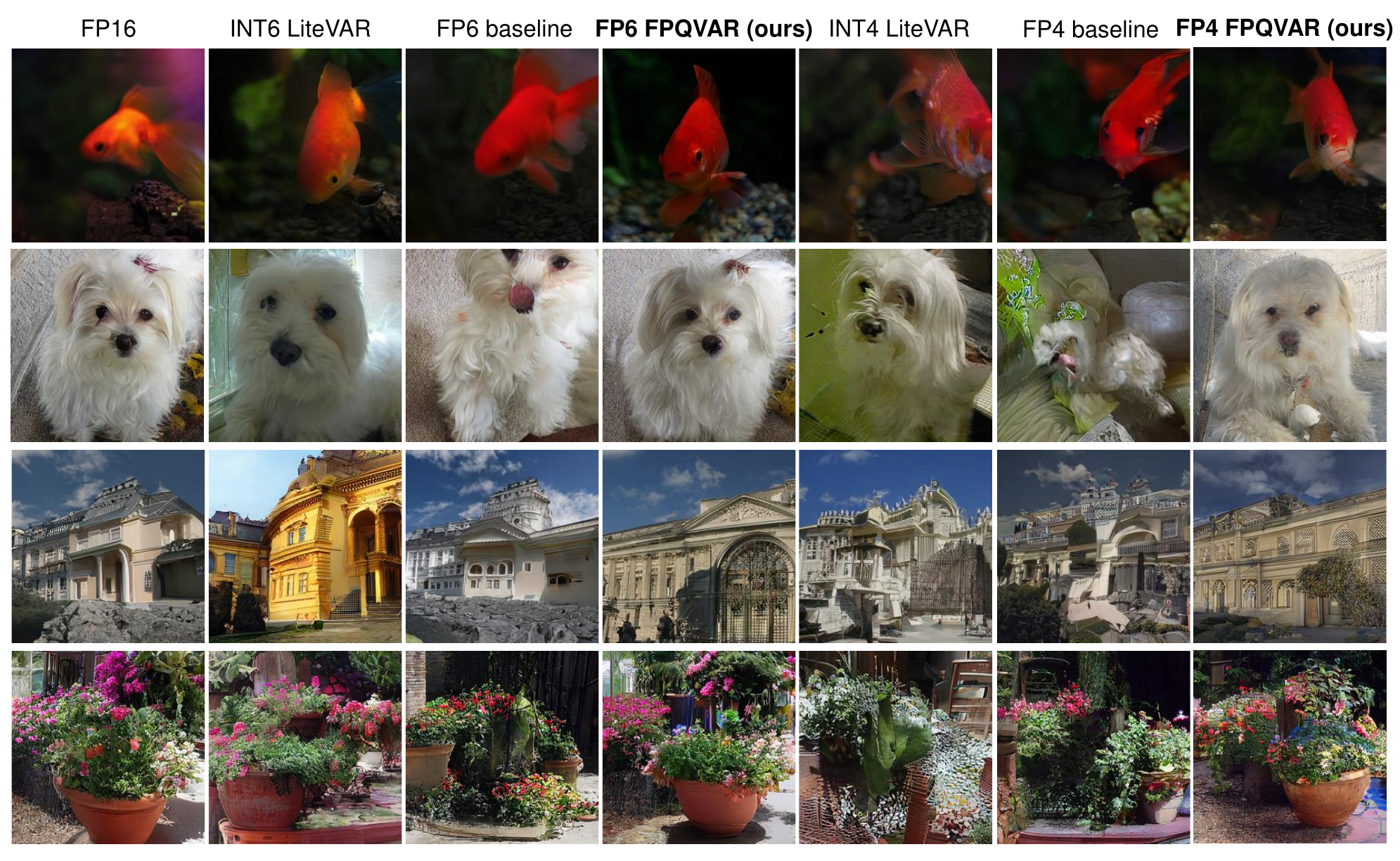}
\caption{Qualitative comparison of $256\times256$ images generated by original FP16 VAR and different quantization methods. The class labels are (from top to bottom): goldfish (1), Maltese dog (153), palace (698), flowerpot (738).}
\label{fig:image_results}
\vspace{-5pt}
\end{figure*}

\begin{figure*}[!tb]
\centering
\includegraphics[width=0.92\linewidth]{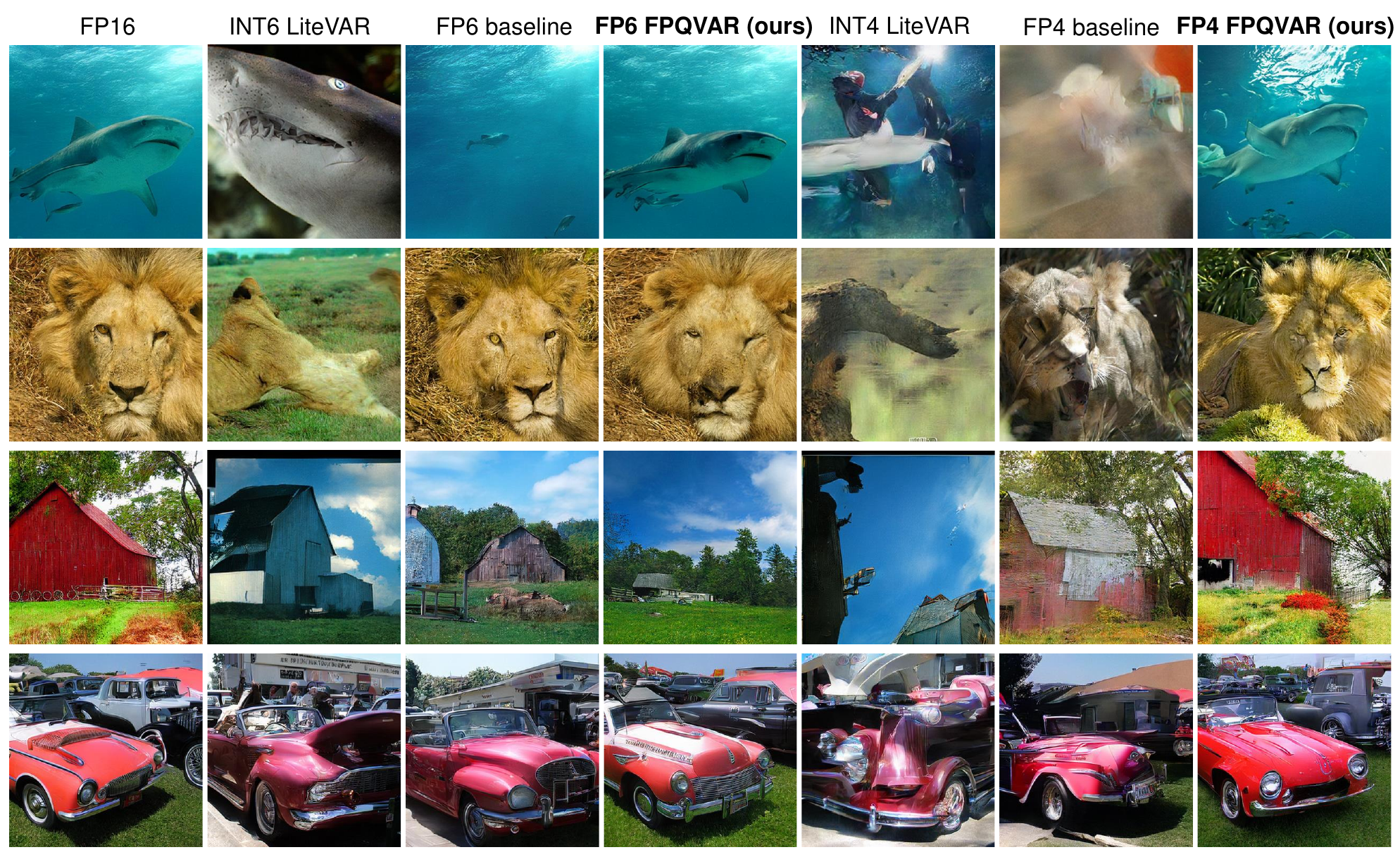}
\caption{Qualitative comparison of $512\times512$ images generated by original FP16 VAR and different quantization methods. The class labels are (from top to bottom): tiger shark (3), lion (291), barn (425), convertible (511).}
\label{fig:image_results_512x512}
\vspace{-5pt}
\end{figure*}

For the qualitative comparison,
we demonstrate some generated $256 \times 256$ and 
$512 \times 512$ images of different methods in Fig.~\ref{fig:image_results}
and Fig.~\ref{fig:image_results_512x512}, respectively.
As can be observed, compared to other methods, 
our proposed FPQVAR generates images with higher quality and fidelity in both 6-bit and 4-bit settings.

\textbf{Hardware Evaluation}
To evaluate our FPQVAR accelerator, we implement an integer-based accelerator with INT W4A4 precision based on LiteVAR~\cite{xie2024litevar} algorithm,
since there is no FPGA-based VAR accelerator so far.
We also report the performance of the original FP16 VAR with FlashAttention~\cite{dao2023flashattention} on an NVIDIA RTX A6000 GPU.
As shown in Table~\ref{tab:hardware comparison},
our FPQVAR accelerator achieves the throughput of 1.1 image/s on VCK190,
which is $3.1\times$ faster than the INT baseline accelerator that performs FP16 computation for FC2 layer.
The energy efficiency of FPQVAR achieves 0.025 image/J, 
which outperforms the INT baseline and GPU baseline by $3.6\times$ and  $2.8\times$, respectively.
The hardware utilization of FPQVAR on VCK190 is listed in Table~\ref{tab:Hardware utilization}.

\begin{table}[!t]
\centering
\caption{Hardware performance comparison.}
\label{tab:hardware comparison}
\resizebox{0.45\textwidth}{!}{
\begin{tabular}{c|c|c|c}
\toprule
& FPQVAR  & INT Baseline & GPU Baseline\\  \midrule

Platform & VCK190 & VCK190   & A6000\\ \midrule

Frequency    & 400 MHz & 400 MHz  & 1.8 GHz \\ \midrule

Precision   & FP4   & INT4   & FP16\\ \midrule

LUT        & 664k  & 697k   & -    \\ 
FF         & 776k   & 852k   & -     \\ 
DSP        & 1840      & 1902   & - \\
BRAM        & 1097       & 1221    &  -    \\
URAM        & 122        & 168    &   -   \\ \midrule

FID $\downarrow$    & 3.58     & 10.83      & 1.98     \\
GOPS             & 6144         & 1966         & 13932    \\
Throughput (img/s)   & 1.1     & 0.36      & 2.6   \\
Energy Eff. (img/J)    & 0.025     & 0.007    & 0.009    \\
\bottomrule
\end{tabular}
}
\end{table}

\begin{table}[!tb]
\centering
\caption{Hardware utilization of FPQVAR on VCK190.}
\label{tab:Hardware utilization}
\scalebox{1.0}{
\begin{tabular}{cc|ccccc}
\toprule
\multicolumn{2}{c}{Modules}             & LUT  & FF   & DSP  & BRAM & URAM \\ \midrule
\multirow{4}{*}{FMU} & Quant1 & 11k  & 11k  & 0    & 0    & 0    \\
                     & GEMM1  & 242k & 204k & 576  & 278  & 32   \\
                     & Quant2 & 19k  & 20k  & 0    & 0    & 0    \\
                     & GEMM2  & 153k & 143k & 384  & 192  & 22   \\  \midrule 
\multicolumn{2}{c|}{SFU}      & 134k & 250k & 824  & 623  & 20   \\ \midrule
\multicolumn{2}{c|}{GHTU}     & 12k & 10k & 56   & 0    & 0    \\ \midrule
\multicolumn{2}{c|}{Others}   & 93k  & 137k & 0    & 4    & 48   \\ \midrule
\multicolumn{2}{c|}{Total}    & 664k & 776k & 1840 & 1097 & 122  \\ 
\bottomrule
\end{tabular}}
\vspace{-5pt}
\end{table}

\subsection{Ablation Study}
\label{subsec:Ablation Study}

\begin{table}[!tb]
\centering
\caption{Ablation study of different components in FPQVAR.}
\label{tab:ablation}
\scalebox{0.9}{
\begin{tabular}{c|cccc}
\toprule
                         & IS $\uparrow$ & FID $\downarrow$  & Precision $\uparrow$ & \begin{tabular}[c]{@{}c@{}}Nomalized\\ Throughput\end{tabular}  \\ \midrule
FP4 baseline      & 167.0          & 10.72     & 0.69       & $1.25\times$      \\
+ DFQ             & 189.5          & 8.25      & 0.72       & $1.02\times$       \\
+ DFQ \& HT        & 232.3          & 4.69     & 0.76       & $0.43\times$     \\
+ DFQ \& GHT       & 219.7          & 5.18     & 0.75       & $1.00\times$      \\
+ DFQ \& GHT \& GALT & 241.5        & 3.58      & 0.77       & $1.00\times$ \\
\bottomrule
\end{tabular}}
\vspace{-5pt}
\end{table}

We study the impact of different components of FPQVAR on model performance
and throughput in Table~\ref{tab:ablation}.
Enabling DFQ significantly enhances the model performance, improving IS from 
167.0 to 189.5 and FID form 10.72 to 8.25.
However, it decreases the throughput from $1.25\times$ to $1.02\times$ due to the introduction of the DFQ GeMM unit,
which already considers our pipeline optimization.
Without our proposed pipeline, the throughput would further decrease to $0.67\times$.
Applying HT on DFQ improves the model performance by reducing the time-varying outliers, 
however, it causes severe throughput degradation from $1.02\times$ to $0.43\times$.
With our proposed GHT and pipeline, the throughput is significantly increased to $1.00\times$.
However, despite being more hardware efficient, GHT is less effective than HT in terms of performance improvement,
since it only rotates the activations within each group.
To address this, we propose GALT, which significantly improves the model performance without degrading the throughput,
which even surpasses the DFQ + HT by 9.2 in IS and 1.11 in FID.

\begin{figure}[!tb]%
  \centering
    \subfloat[Loss curve of QKV projection GALT]{
        \label{subfig:Quantization error of FC2 activation}
        \includegraphics[width=0.24\textwidth]{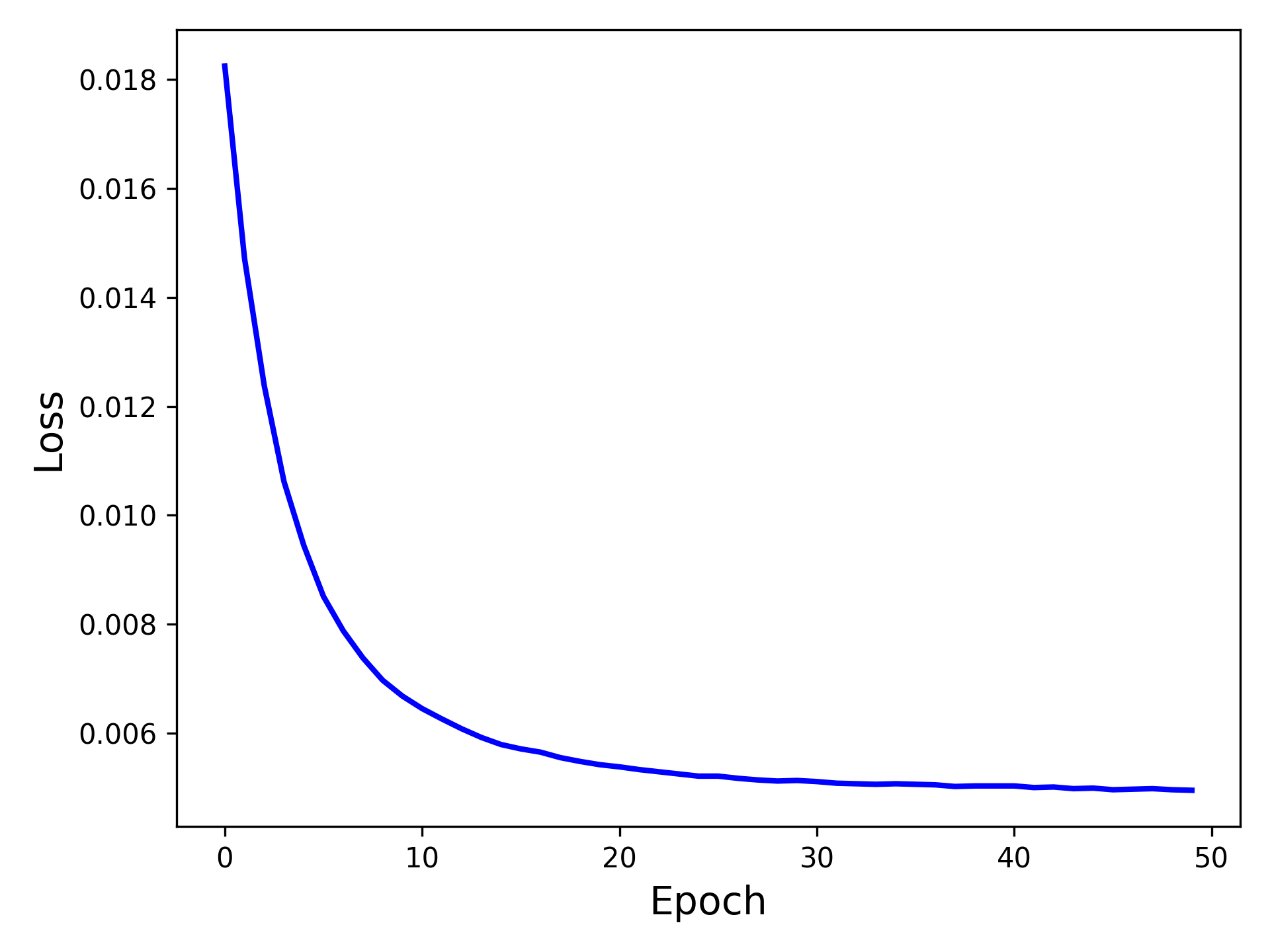}
      }
      \subfloat[Loss curve of FC1 GALT]{
        \label{subfig:Calibration dataset construction}
        \includegraphics[width=0.24\textwidth]{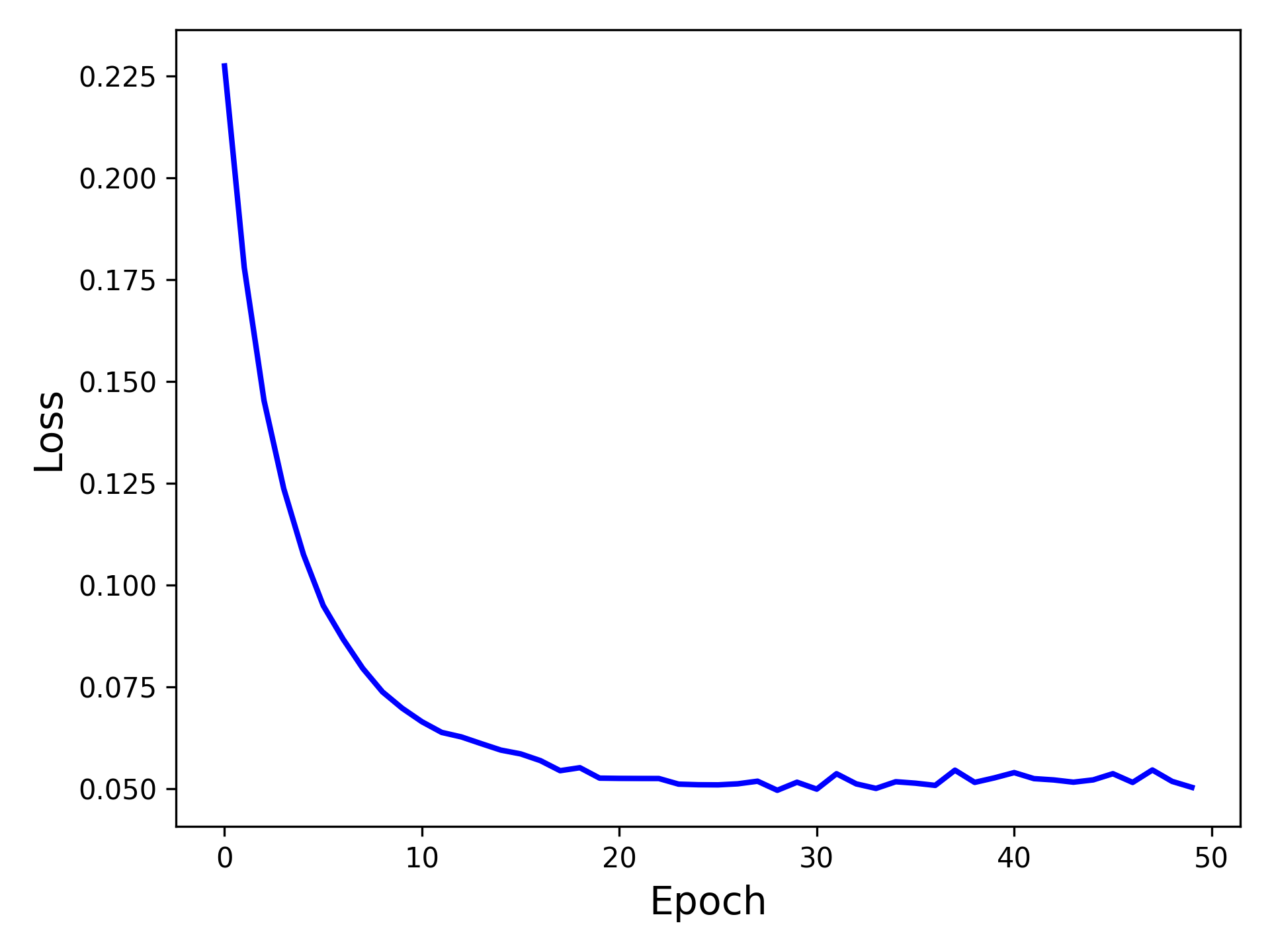}
      }
      
  \caption{
Loss curve of our GALT optimization for the (a) QKV projection layer and (b) FC1 layer.
}
  \label{fig:loss curve}
  \vspace{-5pt}
\end{figure}

We demonstrate the effectiveness of our GALT in Fig.~\ref{fig:loss curve}.
As can be observed, with our proposed optimization algorithm Alg.~\ref{alg: GALT optimize}, the optimization loss, i.e., the quantization error in Eq.~\ref{eq:learnable transformation MSE}
is significantly reduced by $3.6\times$ and $4.6\times$ for QKV projection and FC1 layer, respectively.
Moreover, the optimization speed is also fast,
requiring only 50 minutes for the entire model on a single GPU.

\section{Conclusion}

In this work, we propose FPQVAR, an efficient post-training FP quantization framework for VAR
featuring algorithm and hardware co-design.
First, we identify the challenges of quantizing VAR.
To address these challenges, we propose DFQ for highly imbalanced activation quantization, and propose GHT and GALT to address the time-varying outlier channels.
At the hardware level, we design the first low-bit FP quantizer and multiplier,
and propose the first FPGA-based VAR accelerator featuring low-bit FP inference and an elaborate two-level pipeline.
Extensive experiments show that our FPQVAR significantly improves the model performance compared to
the SOTA method both quantitatively and qualitatively.
Our FPQVAR accelerator on VCK190 achieves a throughput of 1.1 image/s, which is $3.1\times$ faster than the integer-baseline accelerator.
It also demonstrates 3.6× and 2.8× higher energy efficiency
compared to the integer-based accelerator and GPU baseline, respectively.

\bibliographystyle{IEEEtran}
\bibliography{quantization,hardware}

\end{document}